\pdfoutput=1

\documentclass[11pt]{article}

\usepackage[]{ACL2023}

\usepackage{times}
\usepackage{latexsym}

\usepackage[T1]{fontenc}

\usepackage[utf8]{inputenc}

\usepackage{microtype}

\usepackage{inconsolata}

\usepackage{amsfonts} 
\usepackage{amssymb} 
\usepackage{pifont}
\usepackage{tabularx} 
\usepackage{arydshln} 
\usepackage{xcolor}
\usepackage{mathtools} 
\usepackage{adjustbox}
\usepackage{multirow} 
\usepackage{paralist} 
\usepackage{CJKutf8} 
\usepackage{booktabs}

\usepackage{enumerate}
\usepackage{bbding} 
\usepackage{wasysym}
\usepackage{caption}
\usepackage{subcaption}

\usepackage[ruled,linesnumbered]{algorithm2e}
\SetKwInOut{Parameter}{Parameters}

\usepackage{amsthm} 
\theoremstyle{definition}

\usepackage{tikz}
\usepackage[edges]{forest} 
\usepackage{pgfplots}
\usepackage{amsmath}

\usepackage{bm}
\usepackage{scalefnt}

\title{
	MMSD2.0: Towards a Reliable Multi-modal Sarcasm Detection System
}

\author{
	Libo Qin$^{1*}$~~
	Shijue Huang$^{2,4}$\thanks{~~Equally contributed authors.}~~~ 
	Qiguang Chen$^{5}$~~~
	Chenran Cai$^{2,4}$\\
	\textbf{Yudi Zhang}~~~ 
	\textbf{Bin Liang}$^{2,4}$~~~
	\textbf{Wanxiang Che}$^{5}$~~~
    \textbf{Ruifeng Xu}$^{2,3,4}$ \thanks{\,\, Corresponding author.}\\
	$^{1}$School of Computer Science and Engineering, Central South University, China\\
	$^{2}$Harbin Institute of Technology, Shenzhen, China~~~
	$^{3}$Peng Cheng Laboratory, Shenzhen, China\\
	$^{4}$Guangdong Provincial Key Laboratory of Novel Security Intelligence Technologies\\
	$^{5}$Research Center for SCIR, Harbin Institute of Technology, Harbin, China \\
	{\tt lbqin@csu.edu.cn,}~~{\tt \{joehsj310,crcai1023\}@gmail.com,}~~{\tt  yvdi.zhang@outlook.com}\\
{\tt bin.liang@stu.hit.edu.cn,}~~{\tt \{qgchen,car\}@ir.hit.edu.cn,}~~{\tt  xuruifeng@hit.edu.cn}\\
}

\begin{document}
	\maketitle
	
	\begin{abstract}
		Multi-modal sarcasm detection has attracted much recent attention.
		Nevertheless, the existing benchmark (MMSD) has some shortcomings that hinder the development of reliable multi-modal sarcasm detection system:
		(1) There are some spurious cues in MMSD, leading to the model bias learning; (2) The negative samples in MMSD are not always reasonable.
		To solve the aforementioned issues, 
		we introduce 
		MMSD2.0, a correction dataset that fixes the shortcomings of MMSD, by removing the spurious cues and re-annotating the unreasonable samples.
		Meanwhile, we present a novel framework called multi-view CLIP that is capable of leveraging multi-grained cues from multiple perspectives (i.e., text, image, and text-image interaction view) for multi-modal sarcasm detection.
		Extensive experiments show that MMSD2.0 is a valuable benchmark for building reliable multi-modal sarcasm detection systems and multi-view CLIP can significantly outperform the previous best baselines.
	\end{abstract}

\section{Introduction}
\label{Introduction}
\begin{figure}[t]
	\centering
	\centering
	\includegraphics[width=0.48\textwidth]{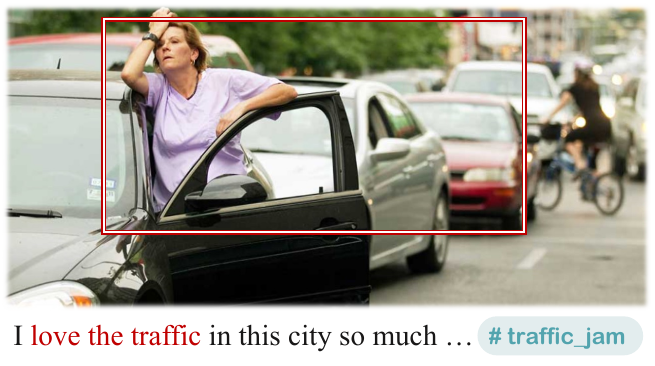}
	\caption{Multi-modal sarcasm example.
		Box and words in red color denote the correlated sarcastic cues. 
		Word \texttt{\#traffic\_jam} with hash symbol \# is a hashtag. 
	}
	\label{fig:intro}
\end{figure}

Sarcasm detection is used to identify the real sentiment of user, which is beneficial for sentiment analysis and opinion mining task~\cite{10.1561/1500000011}.
Recently, due to the rapid progress of social media platform, \textit{multi-modal sarcasm detection}, which aims to recognize the sarcastic sentiment in multi-modal scenario (e.g., text and image modalities), has attracted increasing research attention.
Specifically, as illustrated in \figurename~\ref{fig:intro}, given the text-image pair, multi-modal sarcasm detection system predicts \texttt{sarcasm} label because the image shows a traffic jam that contradicts the text \textit{``love the traffic''}.
Unlike traditional sarcasm detection task,
the uniqueness of multi-modal sarcasm detection lies in effectively modeling the consistency and sarcasm relationship among different modalities.

Thanks to the rapid development
of deep neural networks,
remarkable success has been witnessed in multi-modal sarcasm detection.
Specifically, ~\citet{10.1145/2964284.2964321} makes the first attempt to explicitly concatenate the textual and visual features for multi-modal sarcasm detection.
A series of works have employed attention mechanism to implicitly incorporate features from different modalities~\citep{cai-etal-2019-multi,xu-etal-2020-reasoning,pan-etal-2020-modeling}.
More recently, graph-based approaches have emerged to identify significant cues in sarcasm detection~\citep{10.1145/3474085.3475190, liang-etal-2022-multi,https://doi.org/10.48550/arxiv.2210.03501}, which is capable of better capturing relationship across different modalities, thereby dominating the performance in the literature.

While current multi-modal sarcasm detection systems have achieved promising results, it is unclear whether these results faithfully reflect the multi-modal understanding ability of models. In fact, when a text-modality only model \texttt{RoBERTa} is applied to multimodal sarcasm detection, its performance significantly surpasses the state-of-the-art multi-modal model \texttt{HKE}~\cite{https://doi.org/10.48550/arxiv.2210.03501} with 6.6\% improvement (See Detailed Analysis \S~\ref{sec:MMSD1.0-Performance}).
Such observation suggests that the performance of current models may heavily depend on spurious cues in textual data, rather than truly capturing the relationship among different modalities, resulting in low reliability. Further exploration reveals that the characteristics of the MMSD benchmark~\cite{cai-etal-2019-multi} may be the cause of this phenomenon:
(1) \texttt{Spurious Cues}: 
MMSD benchmark has some spurious cues (e.g., hashtag and emoji word) occuring in an unbalance distribution of positive and negative examples, which leads to the model bias learning; (2) \texttt{Unreasonable Annotation}: MMSD benchmark simply assigns the text without special hashtag (e.g., \#sarcasm, etc.) as negative examples (i.e., not sarcastic label). 
We argue that this construction operation is unreasonable because sentence without \#sarcasm tag can also express the sarcastic intention. Take utterance in \figurename~\ref{fig:overall} as an example, the utterance without  \#sarcasm tag  still belong to the sarcastic sample.
Therefore, further chasing performance on current MMSD benchmark may hinder the development of reliable multi-modal sarcasm detection system.

Motivated by the above observation, we shift our eyes from traditional complex network design work~\citep{cai-etal-2019-multi,10.1145/3474085.3475190, liang-etal-2022-multi} to the establishment of reasonable multi-modal sarcasm detection benchmark.
Specifically, we introduce MMSD2.0 to address these problems.
To solve the first drawback, MMSD2.0 removes the spurious cues (e.g., sarcasm word) from text in the MMSD, which encourages model to truly capture the relationship across different modalities rather than just memorize the spurious correlation.
Such operation can benefit the development of bias mitigation in multi-modal sarcasm detection studies.
To address the second problem, we directly make our efforts to re-annotate the unreasonable data. 
Specifically, given each utterance labeled with ``not sarcastic'' in the MMSD, we use crowdsourced workers to check and re-annotate the label. This correction process results in changes to over 50\% of samples in original MMSD.

In addition to the dataset contribution, we propose a novel framework called \texttt{multi-view CLIP}, which can naturally inherit multi-modal knowledge from pre-trained CLIP model.
Specifically, \texttt{multi-view CLIP} utilizes different sarcasm cues captured from multiple perspectives (i.e., text, image and text-image interaction view), and aggregates multi-view information for final sarcasm detection.
Compared to previous superior graph-based approaches, \texttt{multi-view CLIP} has the following advantages: (1) \texttt{multi-view CLIP} does not require any image pre-processing step for graph building (e.g., object detection);
(2) \texttt{multi-view CLIP} does not require any complex network architecture and can naturally make full use of knowledge in VL pre-trained model for multi-modal sarcasm detection.

Contributions of this work can be summarized as follows:
\begin{itemize}
	\item To the best of our knowledge, we make the first attempt to point out the behind issues in current multi-modal sarcasm benchmark, which motivates researchers to rethink the progress of multi-modal sarcasm detection;
	
	\item We introduce MMSD2.0, a correction dataset removing the spurious cues and fixing unreasonable annotation, for multi-modal sarcasm detection, which takes a meaningful step to build a reliable multi-modal sarcasm system;
	
	\item We propose a novel \texttt{multi-view CLIP} framework to capture different perspectives of image, text and image-text interaction, which attains state-of-the-art performance.

\end{itemize}

To facilitate future research, all datasets, code are publicly available at \url{https://github.com/JoeYing1019/MMSD2.0}.

\section{MMSD2.0 Dataset Construction}
\label{sec:dataset-construction}

\begin{figure*}[t]
	\centering
	\includegraphics[width=\textwidth]{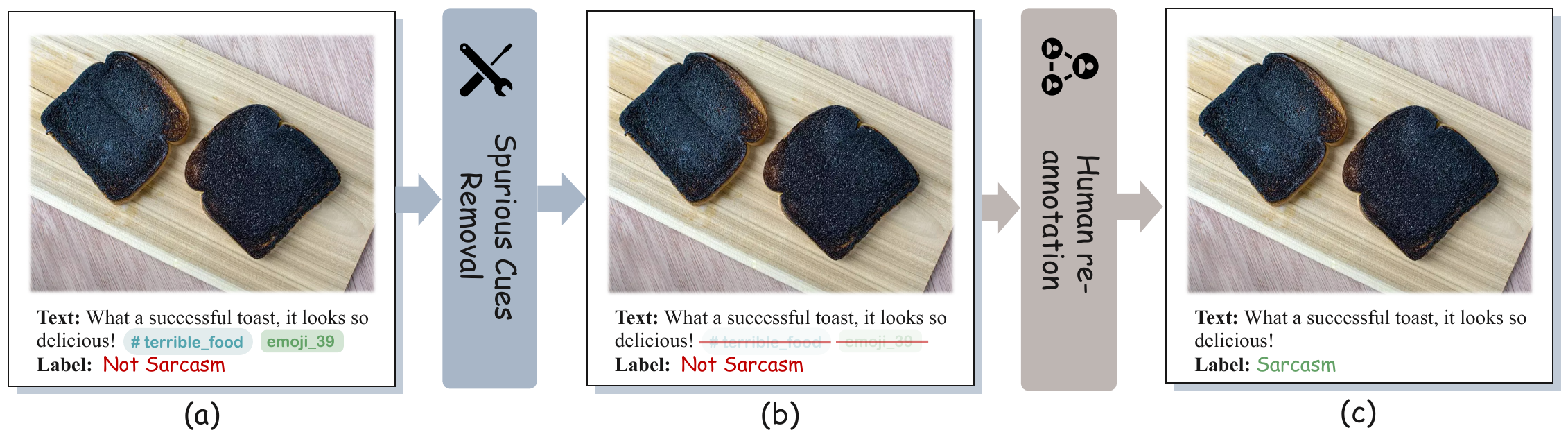}%
	\caption{ Overall process of construction MMSD2.0 dataset.
		Given the example in (a), \texttt{Spurious Cues Removal} stage first remove the spurious cues in text including hashtag word \texttt{(\#terrible\_food)} and emoji word \texttt{(emoji\_39)} to acquire (b), then \texttt{unreasonable samples re-annotation via crowdsourcing} (human re-annotation) stage re-annotates the unreasonable samples to get final reasonable example (c). 
	}
	\label{fig:overall}
\end{figure*}
MMSD2.0 aims to solve two main shortcomings of MMSD, which consists of (1) \texttt{Spurious Cues Removal} ($\S \ref{sec:spurious-removal}$) and (2) \texttt{Unreasonable Samples Re-Annotation via Crowdsourcing} ($\S \ref{sec:human}$).
This section provides MMSD2.0 dataset construction in detail and the overall process is illustrated in \figurename~\ref{fig:overall}.

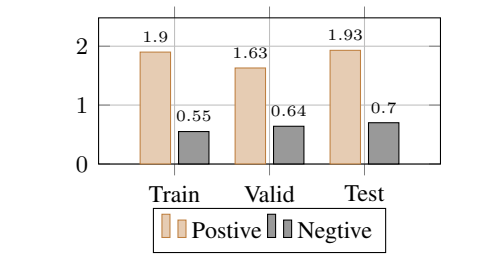
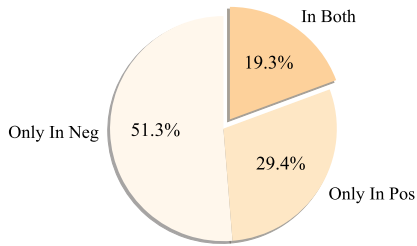
\begin{figure}[t]
	\centering
	\begin{subfigure}{.4\textwidth}
		\centering
		\begin{tikzpicture} 
		\begin{axis}[
		enlargelimits=0.4,
		legend style={at={(0.5,-0.3)},
			anchor=north,legend columns=-1},
		symbolic x coords={Train, Valid, Test },
		xtick=data,
		ybar=3pt,
		bar width=0.4cm,
		width=0.95\linewidth,
		height=.55\linewidth,
		nodes near coords,
		nodes near coords align={vertical},
		nodes near coords style={font=\tiny},
		font=\small,
		grid=major,
		]
		\addplot[fill=brown!40!white,draw=brown] coordinates {
			(Train, 1.90)
			(Valid, 1.63)
			(Test, 1.93)
		};
		\addplot [fill=black!40!white,draw=black] coordinates {
			(Train, 0.55)
			(Valid, 0.64)
			(Test, 0.70)
		};
		\legend{Postive, Negtive}
		\end{axis}
		\end{tikzpicture}
		\caption{Average number of hashtag words per sample in MMSD dataset.}

	\end{subfigure}
	
	\begin{subfigure}{.4\textwidth}
		\centering
		\includegraphics[width=0.85\textwidth]{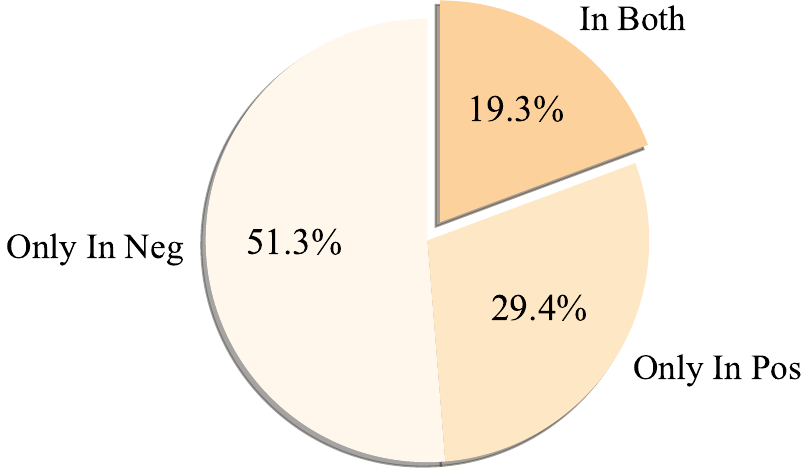}
		\caption{Percentage of emoji words in MMSD dataset.}
		\label{fig:text:sub2}
	\end{subfigure}
	\caption{Text analysis of MMSD dataset.}
	\label{fig:text}
\end{figure}

\subsection{Spurious Cues Removal}\label{sec:spurious-removal}

In our in-depth analysis, we observe that the spurious cues come from two resources: (1) hashtag words and (2) emoji words. 
Therefore, we will remove the two resources, respectively.

\paragraph{Hashtag Word Removal.}
In MMSD, as shown in \figurename~\ref{fig:text}(a), we observe that distribution of hashtag word number in positive sample and negative sample is obviously unbalanced.
As seen, the number of hashtag words in positive sample is on average more than 1 while less than 1 in negative samples in train, validation and test set.
In other word, model only need to learn spurious correlation (hashtag word number) to make a correction prediction rather than truly understand multi-modal correlation in sarcasm detection.

To address the issue, we remove hashtag words from text in the MMSD dataset. This allows the model to capture image features and using them to guide the final prediction, rather than relying on the hashtag word number as a spurious cue. 

\paragraph{Emoji Word Removal.}
Similarly, we find that the distribution of emoji words between positive and negative samples is also unbalanced. 
Specifically, as shown in Figure~\ref{fig:text}(b), 
only 19.3\% of which exist in both positive and negative samples while the rest 80.7\% of emoji words only appear in one type of sample (e.g., positive sample or negative sample). This indicates that model can simply use emoji word distribution as a priority for predicting rather than truly capturing multi-modal cues.

To tackle this issue, we remove all the emoji words in text to force the model learning truly multi-modal sarcasm features rather than relying on spurious textual cues.
 
\subsection{Unreasonable Samples Re-Annotation  via Crowdsourcing}\label{sec:human}
This section consists of two stages for re-annotation: (1) \texttt{sample selection stage} to choose the unreasonable samples and (2) \texttt{re-annotation via crowdsourcing stage} to 
fix the unreasonable samples according to the sample selection stage.
\paragraph{Sample Selection Stage.}
MMSD simply considers the samples without special hashtags like ``\#sarcasm'' as negative sample (i.e., not sarcasm).
In this work, we argue that the process is unreasonable because samples without \#sarcasm tag can also express sarcasm intention.
Therefore, we select all negative samples in the MMSD dataset (over 50\%) as potentially unreasonable samples for further processing.

\paragraph{Re-annotation via Crowdsourcing.}
For all selected samples in  \texttt{sample selection stage}, we 
directly re-annotate them via crowdsourcing by hiring human experts.
Given a sample, each annotator is required to annotate with labels among: 
(i)  \texttt{Sarcasm}: if the sample express sarcasm intention;
(ii) \texttt{Not Sarcasm}: if the sample not express  sarcasm intention;
(iii)  \texttt{Undecided}: if the sample is hard to decide by the annotator.

After the whole annotation is done, the samples with label \texttt{Undecided} is re-annotated again by three experts to decide the final annotation.

\subsection{Quality Control and Data Statistics}
\subsubsection{Quality Control} 
To control quality of the annotated dataset, we conduct two verification methods.
\paragraph{Onboarding Test.}
Before the whole annotation work, we require all the annotators to annotate 100 reasonable samples selected by model and the annotation result will be checked by 3 experts. Only those who achieve 85\% annotation performance can join the annotation process.

\paragraph{Double Check.}
We randomly sample 1,000 annotated samples and ask another new annotator to re-annotate sarcasm label. Then, we calculated the Cohen's Kappa~\cite{McHugh2012InterraterRT} between the previous labels and new labels.
Finally, we get a kappa of 0.811 indicating an almost perfect agreement~\cite{10.2307/2529310}.

\subsubsection{Data Statistics}Table~\ref{tab:data_statistic} provides a detailed comparison of the statistics of the MMSD and MMSD2.0 datasets. We can observe that the distribution of positive and negative samples in MMSD2.0 is more balanced.

\begin{table}[t] 
	\small
	\centering
	\begin{adjustbox}{width=0.5\textwidth}
		\begin{tabular}{l|ccc} 
			\hline
			MMSD/MMSD2.0 & Train & Validation & Test  \\
			\hline
			Sentences & 19,816/19,816 & 2,410/2,410 & 2,409/2,409 \\
			Positive & 8,642/9,572 & 959/1,042 & 959/1,037 \\
			Negative & 11,174/10,240 & 1,451/1,368 & 1,450/1,372 \\
			\hline
		\end{tabular}
	\end{adjustbox}
	\caption{Comparison between MMSD and MMSD2.0. 
	}
	\label{tab:data_statistic}
\end{table}

\section{Approach}
\label{task-description}
This section first describes the basic architecture for \texttt{CLIP} ($\S \ref{sec:CLIP}$) and then illustrates 
the proposed \texttt{multi-view CLIP} framework ($\S \ref{sec:multi-view-clip}$).

\subsection{CLIP}\label{sec:CLIP}
\begin{figure}[t]
	\centering
	\includegraphics[width=0.45\textwidth]{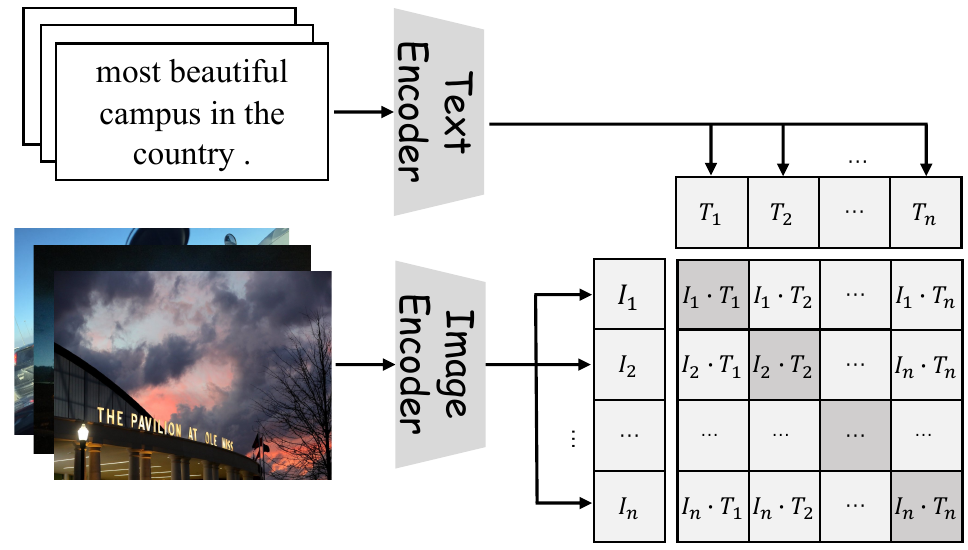}
	\caption{CLIP model.
	}
	\label{fig:CLIP}
\end{figure}

CLIP (Contrastive Language-Image Pretraining)~\cite{pmlr-v139-radford21a}, a powerful vision-and-language pre-trained model, has shown remarkable success on various VL downstream tasks~\cite{9522786, 9897323, https://doi.org/10.48550/arxiv.2205.14304}.
As shown in Figure~\ref{fig:CLIP}
, CLIP contains a visual encoder $\mathbb{V}$ and a text encoder $\mathbb{T}$ where former usually adopts ResNet~\cite{He2015}  or ViT~\cite{dosovitskiy2021an} and the latter employs transformer~\cite{10.5555/3295222.3295349} as backbone. After representation acquired by $\mathbb{T}$ and $\mathbb{V}$, a dot-product function $(\mathbb{T(\cdot)} \cdot \mathbb{V(\cdot)})$ is further used to calculate the similarity of the given image and text and the highest score indicates the corresponding text and image is matched.

\subsection{Multi-View CLIP}\label{sec:multi-view-clip}
\begin{figure*}[t]
	\centering
	\includegraphics[width=0.95\textwidth]{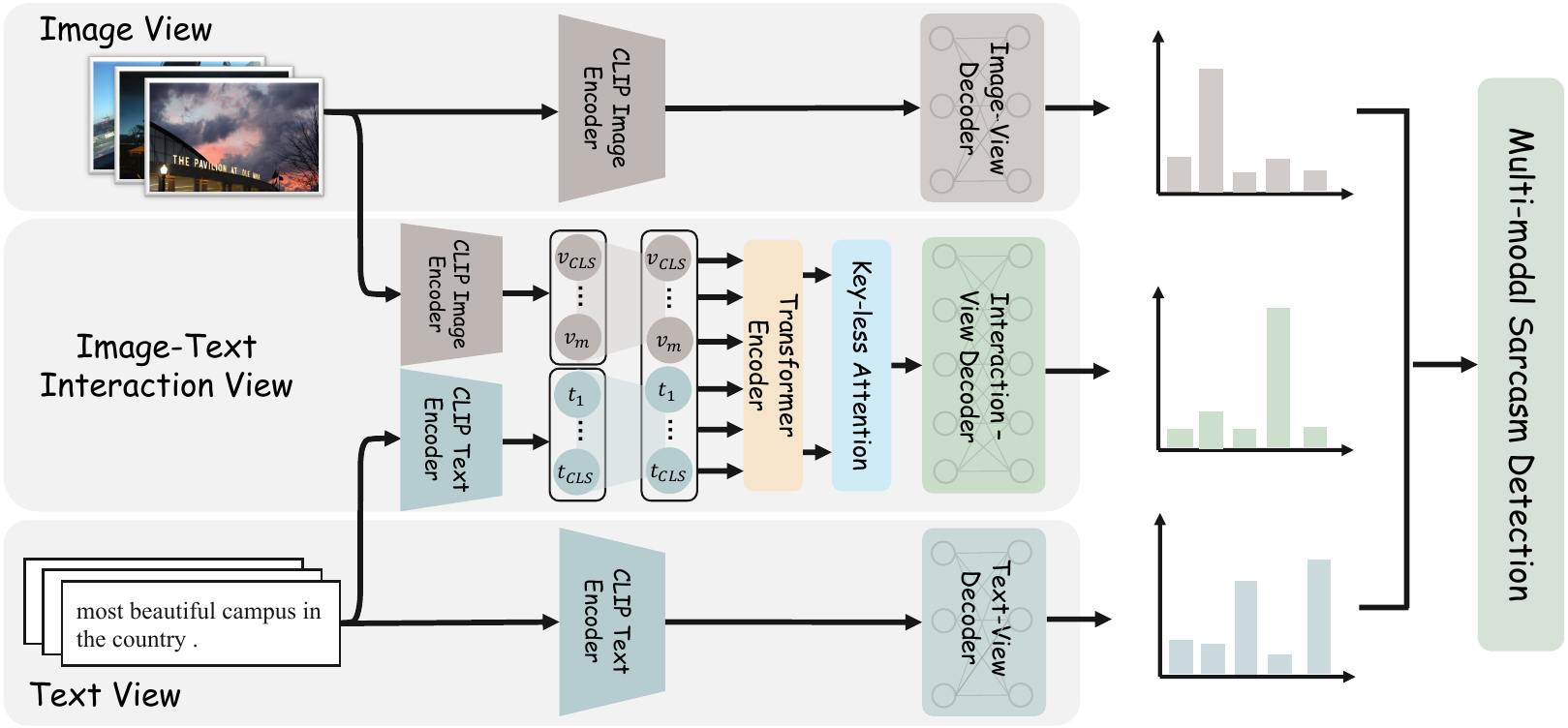}
	\caption{The overall framework of our \texttt{multi-view CLIP}. A pre-trained CLIP model encodes the inputs texts and images. Image view and text view utilize the information of image-only and text-only to capture sarcasm cues. Image-text interaction view fuses the cross-modal information. The three views is aggregated for final prediction.
	}
	\label{fig:CLIP_app}
\end{figure*}

We introduce a novel \texttt{multi-view CLIP} framework for multi-modal sarcasm detection. Our framework consists of text view ($\S \ref{sec:text-clip}$), image view ($\S \ref{sec:image-clip}$) and image-text interaction view ($\S \ref{sec:image-text-clip}$), which explicitly utilize different cues from different views in CLIP model, thereby better capturing rich sarcasm cues for multi-modal sarcasm detection. The overall framework is shown in \figurename~\ref{fig:CLIP_app}.
\subsubsection{Text View} \label{sec:text-clip}
A series of works~\cite{10.1145/3308558.3313735,babanejad-etal-2020-affective}
have shown textual information can be directly used for performing sarcasm detection.
Therefore, we introduce a text view module to judge the sarcasm from text perspective.

Given samples $\mathcal{D}_{} = \left\{(\bm{x}^{(i)},\bm{y}^{(i)})\right\}_{i=1}^{N_{\mathcal{D}_{}}}$ where \{$\bm {x}, \bm{y}\}$ is a text-image pair,  CLIP text encoder $\mathbb T$ is used to output encoding representations $\bm{T}$:
\begin{eqnarray}
	\bm{T} = (\bm{t}_{\text{1}}, \dots,\bm{t}_{\text{n}}, \bm{t}_{\text{CLS}}) = \mathbb {T} (\bm x),
\end{eqnarray}
where $\text{n}$ stands for the sequence length of $\bm x$.

Then, text-view decoder directly employ $\bm{t}_{\text{CLS}}$ for multi-modal sarcasm detection:
\begin{eqnarray}
	\bm y ^{t} = \text{softmax} (\bm{W} \bm{t}_{\text{CLS}} + \bm b),
\end{eqnarray}
where $	\bm y ^{t} $ is output distribution.

\subsubsection{Image View}  \label{sec:image-clip}
Image information can be considered as a textual supplement feature for sarcasm detection~\cite{10.1145/2964284.2964321}
, which motivates us to propose an image CLIP view module to detect  the sarcasm from image perspective.

Specifically, image view CLIP first leverage visual encoder $\mathbb V$ to generate image representation:
\begin{eqnarray}
	\bm{I} = (\bm{v}_{\text{CLS}}, \bm{v}_{\text{1}}, \dots,\bm{v}_{\text{m}}) = \mathbb {V} (\bm y),
\end{eqnarray}
where $\text{m}$ denotes the number of image patches.

Similarly, $\bm{v}_{\text{CLS}}$ is used for prediction:
\begin{eqnarray}
	\bm y ^{v} = \text{softmax} (\bm{W} \bm{v}_{\text{CLS}} + \bm b),
\end{eqnarray}
where $	\bm y ^{v} $ is output distribution.

\subsubsection{Image-text Interaction View}\label{sec:image-text-clip}
Modeling relationship across text and image modality is the key step to multi-modal sarcasm detection.
We follow~\citet{9897323} to use a transformer encoder~\cite{10.5555/3295222.3295349} 
for sufficiently capturing interaction across different modalities.

Specifically, given the yielded text representation $	\bm{T}$ and image representation $	\bm{I}$, we first concat them using $	\bm F $ = $(\bm{v}_{\text{CLS}}, \bm{v}_{\text{1}}, \dots,\bm{v}_{\text{m}}, \bm{t}_{\text{1}}, \dots,\bm{t}_{\text{n}}, \bm{t}_{\text{CLS}})$ = $ \text{Concat} (\bm T, \bm I)$.
Then, we apply different linear functions to obtain the corresponding queries $	\bm Q$, keys $	\bm K$ and values $	\bm V$, and the updated representation $\bm {\hat {F}}$ can be denoted as $	{\bm {\hat{F}}} = \operatorname { softmax } \left( \frac { \bm Q \bm K^\top } { \sqrt { d _ { k } } } \right) \bm V$,
where $d_{k}$ is the mapped dimension.

Given the updated image-text representation  $\bm {\hat {F}}$ = $(\bm{\hat{v}}_{\text{CLS}}, \bm{\hat{v}}_{\text{1}}, \dots,\bm{\hat{v}}_{\text{m}}, \bm{\hat{t}}_{\text{1}}, \dots,\bm{\hat{t}}_{\text{n}}, \bm{\hat{t}}_{\text{CLS}})$, we use keyless attention mechanism~\cite{Long_Gan_Melo_Liu_Li_Li_Wen_2018} to further fuse the image-text interaction feature $\bm f$ by calculating:
\begin{eqnarray}
	\bm f &=& 	\bm p_\text{t}  \bm{\hat{t}}_\text{CLS}  + 	\bm p_\text{v}  \bm{\hat{v}}_\text{CLS},  \\
	\bm p_\text{t}, \bm p_\text{v} &=& \text{softmax} (\bm W (\bm{\hat{t}}_\text{CLS}, \bm{\hat{v}}_\text{CLS}) + \bm b).
\end{eqnarray}

Finally,  $\bm f$ is utilized for prediction:
\begin{eqnarray}
	\bm y ^{f} = \text{softmax} (\bm{W} \bm{f} + \bm b),
\end{eqnarray}
where $	\bm y ^{f} $ is output distribution.

\begin{table*}
	\centering
	\begin{adjustbox}{width=0.95\textwidth}
		\begin{tabular}{l|cccc|cccc}
			\hline
			\multicolumn{1}{c|}{\multirow{2}{*}{Model}} &\multicolumn{4}{c|}{MMSD}&\multicolumn{4}{c}{MMSD2.0}\\
			\cline{2-9}
			&Acc.(\%)&P (\%)&R (\%)&F1 (\%)&Acc. (\%)&P (\%)&R (\%)&F1 (\%)\\
			\hline
			\multicolumn{9}{c}{\textit{Text-modality Methods}} \\
			\hline
			TextCNN~\cite{kim-2014-convolutional}&80.03&74.29&76.39&75.32 & 71.61&64.62&75.22&69.52\\
			Bi-LSTM~\cite{GRAVES2005602}&81.90&76.66&78.42&77.53 & 72.48&68.02&68.08 &68.05\\
			SMSD~\cite{10.1145/3308558.3313735}&80.90&76.46&75.18&75.82 &73.56 &68.45&71.55 &69.97\\
			RoBERTa$^\dagger$~\cite{liu2019roberta}&\textbf{93.97}&\textbf{90.39}&\textbf{94.59}&\textbf{92.45} &79.66 &76.74 &75.70 &76.21\\
			\hline
			\multicolumn{9}{c}{\textit{Image-modality Methods}} \\
			\hline
			ResNet~\cite{He2015}&64.76&54.41&70.80&61.53 &65.50&61.17&54.39&57.58\\
			ViT~\cite{dosovitskiy2021an}&67.83&57.93&70.07&63.40 &72.02 & 65.26& 74.83&69.72\\
			\hline
			\multicolumn{9}{c}{\textit{Multi-Modality Methods}} \\
			\hline
			HFM~\cite{cai-etal-2019-multi}&83.44&76.57&84.15&80.18 &70.57&64.84 &69.05 &66.88\\
			D\&R Net~\cite{xu-etal-2020-reasoning}&84.02&77.97&83.42&80.60 &--&--&--&--\\
			Att-BERT~\cite{pan-etal-2020-modeling}&86.05&80.87&85.08&82.92 &80.03 &76.28 &77.82 &77.04\\
			InCrossMGs~\cite{10.1145/3474085.3475190}&86.10&81.38&84.36&82.84 &--&--&--&--\\
			CMGCN~\cite{liang-etal-2022-multi}&86.54&--&--&82.73 &79.83 &75.82 &78.01 &76.90\\
			HKE~\cite{https://doi.org/10.48550/arxiv.2210.03501}&87.36&81.84&86.48&84.09 &76.50 &73.48 &71.07 &72.25\\
			\cdashline{1-9}[1pt/1pt]
			Multi-view CLIP&\textbf{88.33}&\textbf{82.66}&\textbf{88.65}&\textbf{85.55}& \textbf{85.64} &\textbf{80.33} &\textbf{88.24} &\textbf{84.10}\\
			\hline
		\end{tabular}
	\end{adjustbox}
	\caption{
		Main Results. Baseline results on dataset MMSD are taken from~\citet{https://doi.org/10.48550/arxiv.2210.03501}. Results with - denote
		the code is not released.
		Results with $\dagger$ stand for that we re-implement the model. 
	}
	\label{exp:main_results}
\end{table*}
\subsection{Multi-view Aggregation}\label{sec:multi-view-aggregation}
Given the obtained $\bm y^{t}, \bm y^{v}, \bm y^{f}$, we adopt a late fusion ~\cite{10.1109/TPAMI.2018.2798607}
to yield the final prediction $\bm y ^{o} $:
\begin{eqnarray}
	\bm y ^{o} = \bm y^{t} + \bm y^{v} + \bm y^{f},
\end{eqnarray}
where $\bm y ^{o}$ can be regarded as leveraging rich features from different perspectives of text view, image view and image-text interaction view.

\subsection{Model Training}\label{sec:model-training}
We use a standard binary entropy loss for image, text, and image-text interaction view, and a joint optimization for training the entire framework:

\begin{small} 
	\begin{eqnarray}
		\mathcal{L} =\sum_{i \in \{\text{t}, \text{v}, \text{f}\} } (\bm {\hat{y}^{\text{i}}}\cdot \log(\bm y^{\text{i}}) + (1-\bm {\hat{y}^{\text{i}}})\log(1-(\bm y^{\text{i}})),
	\end{eqnarray}
\end{small}
where $\bm {\hat{y}^{\text{i}}}$ is the gold label. 
It is worth noting that we can directly use the final gold label for training image-view CLIP and text-view CLIP, which does not bring any annotation burden.

\section{Experiments}
\label{experiments}
\subsection{Experimental Settings}
We conduct experiments on MMSD~\citep{cai-etal-2019-multi} and MMSD2.0. 
We implement our model based on the Huggingface Library~\cite{wolf2020huggingfaces} and adopt \texttt{clip-vit-base-patch32} as backbone. We use AdamW~\cite{loshchilov2018decoupled} as optimizer to optimize the parameters in our model. The batch size and epoch for training are 32 and 10, respectively. The learning rate for CLIP is $1e^{\text{-}6}$ and for the other parts is $5e^{\text{-}4}$. 
We select all hyper-parameters from the validation set. All experiments are conducted at Tesla V100s.

\subsection{Baselines}
Following~\citet{https://doi.org/10.48550/arxiv.2210.03501}, we explore three type of baselines: (i) \textit{text-modality methods}, (ii) \textit{image-modality methods} and (iii) \textit{multi-modality methods}.

(i) For \textit{text-modality methods}, we employ 
(1) \texttt{TextCNN}~\cite{kim-2014-convolutional} 
(2) \texttt{Bi-LSTM}~\cite{GRAVES2005602} 
(3) \texttt{SMSD}~\cite{10.1145/3308558.3313735} 
and (4) \texttt{RoBERTa}~\cite{liu2019roberta}.

(ii) For \textit{image-modality methods}, our exploration including: (1) \texttt{ResNet}~\cite{He2015} and (2)  \texttt{ViT}~\cite{dosovitskiy2021an}, which are widely used backbone in CLIP.

(iii) For \textit{multi-modality methods}, we compare \texttt{multi-view CLIP} with the following state-of-the-art baselines:
(1) \texttt{HFM}~\cite{cai-etal-2019-multi} is a hierarchical fusion model for multi-modal sarcasm detection;
(2) \texttt{D\&R Net}~\cite{xu-etal-2020-reasoning} proposes a decomposition and relation network to model cross-modality feature;
(3) \texttt{Att-BERT}~\cite{pan-etal-2020-modeling} applies two attention mechanisms to  model the text-only and cross-modal incongruity, respectively;
(4) \texttt{InCrossMGs}~\cite{10.1145/3474085.3475190} is a graph-based model using in-modal and cross-modal graphs to capture sarcasm cues;
(5) \texttt{CMGCN}~\cite{liang-etal-2022-multi} proposes a fine-grained cross-modal graph architecture to model the cross modality information;
(6) \texttt{HKE}~\cite{https://doi.org/10.48550/arxiv.2210.03501} is a hierarchical graph-based framework to model atomic-level and composition-level congruity. 
For a fair comparison, \texttt{CMGCN} and \texttt{HKE} are the versions without external knowledge.

\subsection{Performance on MMSD}\label{sec:MMSD1.0-Performance}
Following ~\citet{https://doi.org/10.48550/arxiv.2210.03501}, we adopt accuracy (Acc.), F1 score (F1), precision (P) and recall (R) to evaluate the model performance. 

Table~\ref{exp:main_results} (left part) illustrates the results on MMSD. We have the following observations: (1) \textit{Text-modality methods} achieve promising performance and \texttt{RoBERTa} even surpasses multi-modality approaches, which indicates the sarcasm detection model can only rely on the text features to make a correct prediction, supporting the motivation of our work;
(2) In \textit{multi-modality approaches}, \texttt{multi-view CLIP} attains the best results, demonstrating the effectiveness of integrating features from different modality views.

\subsection{Performance on MMSD2.0}\label{sec:MMSD2.0-Performance}
Table~\ref{exp:main_results} (right part) presents the results on MMSD2.0.
As seen, we have the following observations:
(1) Compared with the MMSD dataset, performance of all the baselines has decreased in different degrees on MMSD2.0 dataset. Especially the single-modality model \texttt{RoBERTa}'s performance decrease by 16.24\% on F1 score. This indicates that MMSD2.0 successfully removes the spurious cues and can be used for a more reliable benchmark;
(2) Additionally,  the previous graph-based baselines such as \texttt{CMGCN} and \texttt{HKE} do not make good performance on MMSD2.0, even worse than the \texttt{Att-BERT}.
We attribute it to the fact that such graph-based approaches highly rely on the spurious cues (e.g., hashtag and emoji word) when constructing the text semantic dependency graph, which cannot be accessed in MMSD2.0;
(3) Lastly, \texttt{multi-view CLIP} not only attains the best results in multi-modality approaches but also in both text and image modality approaches, which further verifies the effectiveness of our framework.

\subsection{Analysis}
To understand the \texttt{multi-view CLIP} in more depth, we answer the following research questions:
(1) Does each modality view contribute to the overall performance of the model?
(2) What effect do the training strategies of CLIP have on the model performance?
(3) What impact do different interaction approaches have on the model performance?
(4) Does the \texttt{multi-view CLIP} approach remain effective in low-resource scenarios?
(5) Why does the \texttt{multi-view CLIP} work?

\subsubsection{Answer1: All Views Make Contribute To Final Performance}
\begin{table}[t]
	\centering
	\begin{adjustbox}{width=0.4\textwidth}
		\begin{tabular}{c|cccc}
			\hline
			Model&Acc.(\%)&P(\%)&R(\%)&F1(\%)\\
			\hline
			Ours&\textbf{85.64} &80.33 &88.24&\textbf{84.10}\\
			w/o $\mathcal{L_T}$&84.18&80.60&83.32&81.93\\
			w/o $\mathcal{L_V}$ &83.69&76.97&88.62&82.38\\
			w/o $\mathcal{L_F}$&82.44&73.80&91.80&81.82\\
			\hline
		\end{tabular}
	\end{adjustbox}
	\caption{Ablation study of \texttt{multi-view CLIP} on MMSD2.0.}
	\label{exp:abulation}
\end{table}

To analyze the effect of different modality views of our framework, we remove the training object of text view $\mathcal{L_T}$, image view $\mathcal{L_V}$ and image-text interaction view $\mathcal{L_F}$ separately.

Table~\ref{exp:abulation} illustrates the results. We can observe the accuracy decrease by 1.46\%, 1.95\% and 3.20\% when removing  $\mathcal{L_T}$, $\mathcal{L_V}$ and $\mathcal{L_F}$, respectively.
As seen, our framework attains the best performance when combining all these views.  This suggests that all views in our framework make contribute to the final performance.
It is worth noting that the removal of $\mathcal{L_F}$ leads to a significant performance drop, which suggests that modeling the multi-modality interaction features is crucial for multi-modal sarcasm detection.

\subsubsection{Answer2: Full Finetuned Gains The Best Performance}
\begin{figure}[t]
	\centering
	\resizebox{0.8\columnwidth}{!}{%
		\begin{tikzpicture}
		\begin{axis} [
		ybar,
		enlargelimits=0.15,
		bar width=0.4cm,
		ybar=4pt,
		width=0.95\linewidth,
		height=.5\linewidth,
		ymin = 80, 
		ymax = 89,
		grid=major,
		xticklabel style = {font=\small,yshift=1.0ex},
		symbolic x coords={
			Freeze All, Freeze VE, Freeze TE, Full Finetuned
		},
		legend style={
			at={(0,1.0)},
			anchor=north west,
			legend columns=-1
		},
		font=\small,
		xtick=data,
		x tick label style={rotate=10},
		nodes near coords,
		nodes near coords align={vertical},
		every node near coord/.append style={font=\tiny},
		]
		\addplot[fill=brown!40!white,draw=brown] coordinates {
			(Freeze All, 84.72) (Freeze VE, 84.85) (Freeze TE, 84.93) (Full Finetuned, 85.64) 
		};
		\addplot[fill=black!40!white,draw=black] coordinates {
			(Freeze All, 83.64) (Freeze VE, 83.60) (Freeze TE, 83.48) (Full Finetuned, 84.10) 
		};
		\legend{Acc.(\%), F1(\%)}
		\end{axis}
		\end{tikzpicture}}
	\caption{Different fine-tuning methods. Freeze All stands for both $\mathbb{V}$ and $\mathbb{T}$ are frozen, Freeze VE and Freeze TE indicate $\mathbb{V}$  and $\mathbb{T}$ are frozen respectively, and Full Finetuned means the whole CLIP is trainable.
	}
	\label{exp:freeze}
\end{figure}
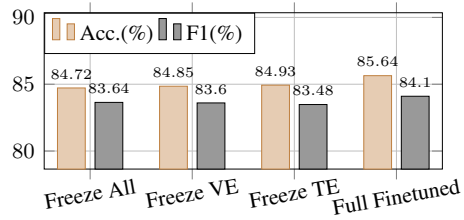

To investigate the influence of training  methods of backbone network CLIP, we 
conduct experiments on different training combination of $\mathbb{V}$ and $\mathbb{T}$.

The results are shown in \figurename~\ref{exp:freeze}.
It reveals that full finetuning of CLIP leads to the best performance. An interesting observation is that freezing all of CLIP leads to almost the same performance as freezing only $\mathbb{T}$ or $\mathbb{V}$. This can be attributed to the fact that the text and image representations in CLIP are aligned, and only training a single part can break this alignment property.

\subsubsection{Answer3: Transformer Interaction Fuse Cross-modal Information Deeper}
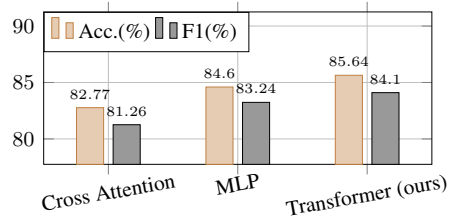
\begin{figure}[t]
	\centering
	\resizebox{0.8\columnwidth}{!}{
		\begin{tikzpicture}
		\begin{axis} [
		ybar,
		enlargelimits=0.25,
		bar width=0.4cm,
		width=0.95\linewidth,
		height=.5\linewidth,
		ymin = 80, 
		ymax = 89,
		ybar=4pt,
		grid=major,
		xticklabel style = {font=\small,yshift=1.0ex},
		symbolic x coords={
			Cross Attention, MLP, Transformer (ours)
		},
		legend style={
			at={(0,1.0)},
			anchor=north west,
			legend columns=-1
		},
		font=\small,
		xtick=data,
		x tick label style={rotate=10},
		nodes near coords,
		nodes near coords align={vertical},
		every node near coord/.append style={font=\tiny},
		]
		\addplot[fill=brown!40!white,draw=brown] coordinates {
			(Cross Attention, 82.77) (MLP, 84.60) (Transformer (ours), 85.64)
		};
		\addplot[fill=black!40!white,draw=black] coordinates {
			(Cross Attention, 81.26) (MLP, 83.24) (Transformer (ours), 84.10)
		};
		\legend{Acc.(\%), F1(\%)}
		\end{axis}
		\end{tikzpicture}
	}
	\caption{Different interaction methods. Cross attention is based on self-attention but obtains query from one modality and key, value from another modality. MLP denotes applying a feed-forward layer after concatenating the information from different modalities.
	}
	\label{exp:interact}
\end{figure}
To further verify the effectiveness of our framework, we explore different interaction methods in text-image interaction view, including cross attention~\cite{pan-etal-2020-modeling} and the MLP approach which concatenates information from different modalities and uses feedforward layer to fuse them.

\figurename~\ref{exp:interact} illustrates the results. We can observe that our transformer interaction is more efficient than other two interactions.
We attribute it to the fact that the transformer interaction is able to fuse the information from different modals in a more depth way, hence attaining the best performance.

\subsubsection{Answer4: Multi-view CLIP can Transfer to Low-resource Scenario}
\begin{figure}[t]
	\centering
	\resizebox{0.87\columnwidth}{!}{
		\begin{tikzpicture}
		\begin{axis}[
		ylabel=Acc. (\%),
		width=7cm, height=4.2cm,
		ymin=70, ymax=89,
		xtick={1, 2, 3, 4},
		xticklabels={10\%, 20\%, 50\%, 100\%},
		xlabel near ticks,
		ylabel near ticks,
		xtick pos=left,
		ytick pos=left,
		ytick align=inside,
		xtick align=inside,
		ymajorgrids=true,
		grid style=dashed,
		legend pos=south east,
		legend style={nodes={scale=0.8, transform shape}}, 
		font=\small,
		]
		\addplot+[semithick,mark=*,mark options={scale=0.7}, brown!50!white] plot coordinates {
			(1, 81.20)
			(2, 81.69)
			(3, 84.43)
			(4, 85.64)
		};
		\addplot+[semithick,mark=x,mark options={scale=0.7}, black!40!white] plot coordinates {
			(1, 73.06)
			(2, 76.34)
			(3, 79.12)
			(4, 80.03)
		};
		\legend{Multi-vew CLP, Att-BERT}
		\end{axis}
		\end{tikzpicture}
	}
	\caption{Low-Resource Performance.
	}
	\label{exp:low_resources}
\end{figure}
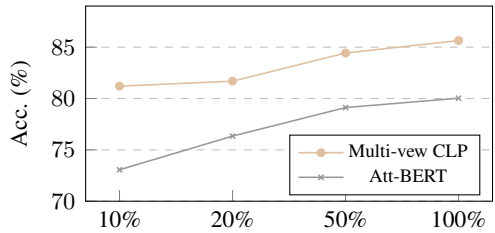

To explore the effectiveness of \texttt{multi-view CLIP} in low-resource scenario, we experiment with different training sizes including 10\%, 20\% and 50\%.

The results are shown in Figure~\ref{exp:low_resources}, indicating that the \texttt{multi-view CLIP} approach still outperforms other baselines in low-resource settings. 
In particular, when the data size is 10\%, the \texttt{multi-view CLIP} approach outperforms the baselines by a large margin in the same setting, and even surpasses \texttt{Att-BERT} when trained on the full dataset.
We attribute this to the fact that the knowledge learned during the CLIP pre-training stage can be transferred to low-resource settings, indicating that our framework is able to extract sarcasm cues even when the training corpus is limited in size.

\subsubsection{Answer5: Multi-view CLIP Utilize Correct Cues}

To further explain why our framework works, we visualize the attention distribution of visual encoder $\mathbb V$ to show why our framework is effective for multi-modal sarcasm detection. 

\begin{figure}[t]
	\centering
	\centering
	\includegraphics[width=0.45\textwidth]{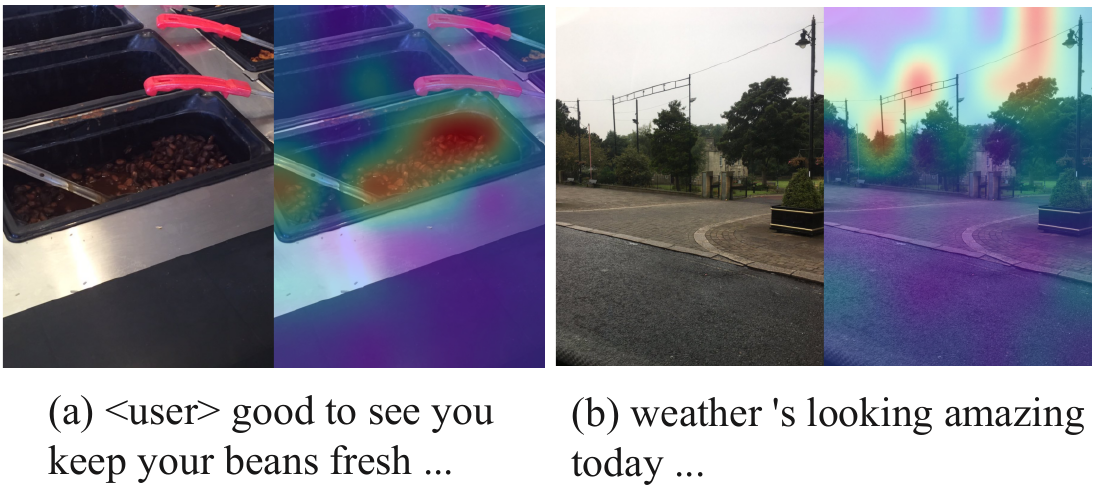}
	\caption{Visualization.
	}
	\label{exp:attention}
\end{figure}

As shown in \figurename~\ref{exp:attention}, our model can successfully focus on appropriate parts of the image with sarcasm cues. For example, our framework pays more attention on the no-fresh beans in \figurename~\ref{exp:attention}(a), which as the important cues to contradicted with \textit{``fresh''} in text. Meanwhile, our framework focus on the bad weather regions in \figurename~\ref{exp:attention}(b), which also an important cues incongruent with \textit{``amazing''} in text. 
This demonstrates that our framework can successfully focus on the correct cues.

\section{Related Work}
\label{related-work}

Sarcasm detection identifies the incongruity of sentiment from the context, which first attracts attention in text modality.
Early studies focused on using feature engineering approaches to detect incongruity in text~\cite{6761575,Bamman2015ContextualizedSD}.
A series of works~\citep{poria-etal-2016-deeper,zhang-etal-2016-tweet,10.1145/3308558.3313735} explored deep learning network (e.g., CNN, LSTM and self-attention) for sarcasm detection. 
\citet{babanejad-etal-2020-affective}  extended  BERT by
incorporating into it effective features for sarcasm detection.
Compared to their work, we aim to solve sarcasm detection in the multi-modal scenario while their work focus on text sarcasm detection.

As the rapid popularization of social medial platform, multi-modal sarcasm detection attracts increasing research attention in recent years. 
~\citet{10.1145/2964284.2964321}  explored the multi-modal sarcasm detection task for the first time and tackled this task by concatenating the features from text and image modalities respectively.
~\citet{cai-etal-2019-multi} proposed a hierarchical fusion model to fuse the information among different modalities and resealed a new public dataset.
~\citet{xu-etal-2020-reasoning} suggested to represent the commonality and discrepancy between image and text by a decomposition and relation network.
~\citet{pan-etal-2020-modeling} implied a BERT architecture-based model to consider the incongruity character of sarcasm.
~\citet{10.1145/3474085.3475190} explored constructing interactive in-modal and cross-modal  to learn sarcastic features.
~\citet{liang-etal-2022-multi} proposed a cross-modal graph architecture to model the fine-grained relationship between text and image modalities.
~\citet{https://doi.org/10.48550/arxiv.2210.03501} proposed a hierarchical framework to model both atomic-level congruity and composition-level congruity.
In contrast to their work, we make the first attempt to address the bias issues in traditional MMSD dataset, towards to building a reliable multi-modal sarcasm detection system. In addition, we introduce MMSD2.0 to this end, aiming to facilitate the research.
To the best of our knowledge, this is the first work to reveal the spurious cues in current multi-modal sarcasm detection dataset.

\section{Conclusion}
\label{conclusion}
This paper first analyzed the underlying issues in current multi-modal sarcasm detection dataset and then introduced a 
MMSD2.0 benchmark, which takes the first meaningful step towards building a reliable multi-modal sarcasm detection system.
Furthermore, we proposed a novel framework \texttt{multi-view CLIP} to capture sarcasm cues from different perspectives, including image view, text view, and image-text interactions view. Experimental results show that \texttt{multi-view CLIP} attains state-of-the-art performance.

	\section*{Limitations}
	This work contributes a debias benchmark MMSD2.0 for building reliable multi-modal sarcasm detection system. While appealing, MMSD2.0 is built on the available MMSD benchmark. In the future, we can consider annotating more data to break through the scale and diversity of the original MMSD.
	
	\section*{Ethics Statement}
	All data of MMSD2.0 come from the MMSD dataset~\citet{cai-etal-2019-multi}, which is an open-source dataset available for academic research. 
	Our annotation process was carried out by annotators who were postgraduate students at Chinese universities and they were paid properly.

	\section*{Acknowledgments}
	This work was partially supported by the National Natural Science Foundation of China (62006062, 62176076, 62236004, 61976072), National Key R\&D Program of China via grant 2020AAA0106501, Natural Science Foundation of GuangDong 2023A1515012922, the Shenzhen Foundational Research Funding (JCYJ20220818102415032), the Major Key Project of PCL2021A06, Guangdong Provincial Key Labo-ratory of Novel Security Intelligence Technologies 2022B1212010005.

	\bibliography{anthology,custom}
	\bibliographystyle{acl_natbib}
	
\end{document}